\documentclass[review]{elsarticle}

\usepackage{algorithm}
\usepackage{algpseudocode}
\usepackage{wrapfig}
\usepackage{amsmath}
\usepackage{amsfonts}
\usepackage{booktabs}
\usepackage{graphicx}

\usepackage{lineno,hyperref}
\modulolinenumbers[5]

\journal{Journal of Computer and Systems Sciences}

\begin{document}

\begin{frontmatter}

\title{Spatially Encoding Temporal Correlations to Classify Temporal Data Using Convolutional Neural Networks
\tnoteref{mytitlenote}}
\tnotetext[mytitlenote]{Preliminary versions of parts of this paper appear in the Twenty-Ninth AAAI workshop proceedings on Trajectory-based Behaviour Analytics.}

\author[mymainaddress]{Zhiguang Wang\corref{mycorrespondingauthor}}
\cortext[mycorrespondingauthor]{Corresponding author.}
\ead{zgwang813@gmail.com}

\author[mymainaddress]{Tim Oates}

\address[mymainaddress]{Department of Computer Science and Electric Engineering, University of Maryland Baltimore County, Baltimore, 21228 Maryland, United States}

\begin{abstract}
We propose an off-line approach to explicitly encode temporal patterns spatially as different types of images, namely, Gramian Angular Fields and Markov Transition Fields. This enables the use of
techniques from computer vision for feature learning and classification. We used Tiled Convolutional Neural Networks to learn
high-level features from individual GAF, MTF, and GAF-MTF images on 12 benchmark time series datasets and two real spatial-temporal trajectory datasets. The classification results of our approach are competitive with
state-of-the-art approaches on both types of data. An analysis of the features and
weights learned by the CNNs explains why the approach works.
\end{abstract}

\begin{keyword}
Time-series\sep Trajectory \sep Classification \sep Gramian Angular Field\sep Markov Transition Field \sep Convolutional Neural Networks
\end{keyword}

\end{frontmatter}


\section{Introduction}
The problem of temporal data classification has attracted great
interest recently, finding applications in domains as diverse
as medicine, finance, entertainment, and industry. However, learning the complicated temporal correlations in complex dynamic systems is still a challenging problem. Inspired by recent successes of deep learning in computer vision, we consider the problem of encoding temporal information spatially as images to
allow machines to ''visually'' recognize and classify temporal data, especially time series data. 

Recognition tasks in speech and audio have
been well studied. Researchers have achieved success using
combinations of HMMs with acoustic models based on Gaussian Mixture
models (GMMs) \cite{reynolds1995robust,leggetter1995maximum}. An
alternative approach is to use deep neural networks to produce 
posterior probabilities over HMM states. Deep learning has become
increasingly popular since the introduction of effective ways to train
multiple hidden layers \cite{hinton2006fast} and has been proposed as
a replacement for GMMs to model acoustic data in speech recognition
tasks \cite{mohamed2012acoustic}. These Deep Neural Network - Hidden
Markov Model hybrid systems (DNN-HMM) achieved remarkable performance
in a variety of speech recognition tasks
\cite{hinton2012deep,deng2013new,deng2013recent}. Such success stems
from learning distributed representations via deeply layered structure
and unsupervised pretraining by stacking single layer Restricted
Boltzmann Machines (RBM).

Another deep learning architecture used in computer vision is
convolutional neural networks (CNNs) \cite{lecun1998gradient}.  CNNs
exploit translational invariance within their structures by extracting
features through receptive fields \cite{hubel1962receptive} and learn
with weight sharing. CNNs are the state-of-the-art approach in various
image recognition and computer vision tasks
\cite{lawrence1997face,krizhevsky2012imagenet,lecun2010convolutional}. Since
unsupervised pretraining has been shown to improve performance
\cite{erhan2010does}, sparse coding and Topographic Independent
Component Analysis (TICA) are integrated as unsupervised pretraining
approaches to learn more diverse features with complex invariances
\cite{kavukcuoglu2010learning,ngiam2010tiled}.

CNNs were proposed for speech processing because of their invariance to
shifts in time and frequency
\cite{lecun1995convolutional}. Recently, CNNs have been shown to
further improve hybrid model performance by applying convolution and
max-pooling in the frequency domain on the TIMIT phone recognition
task \cite{abdel2012applying}. A heterogeneous pooling approach proved
to be beneficial for training acoustic invariance
\cite{deng2013deep}. Further exploration with limited weight sharing
and a weighted softmax pooling layer has been proposed to optimize CNN
structures for speech recognition tasks \cite{abdel2013exploring}.

However, except for audio and speech data, relatively little work has explored
feature learning in the context of typical time series analysis tasks
with current deep learning architectures. \cite{zheng2014time}
explores supervised feature learning with CNNs to classify
multi-channel time series with two datasets. They extracted
subsequences with sliding windows and compared their results to
Dynamic Time Warping (DTW) with a 1-Nearest-Neighbor classifier
(1NN-DTW). Our motivation is to explore a novel framework to encode
time series as images and thus to take advantage of the success of
deep learning architectures in computer vision to learn features and
identify structure in time series. Unlike speech recognition systems
in which acoustic/speech data input is typically represented by
concatenating Mel-frequency cepstral coefficients (MFCCs) or
perceptual linear predictive coefficient (PLPs)
\cite{hermansky1990perceptual}, typical time series data are not
likely to benefit from transformations applied to speech or
acoustic data.

In this work, we propose two types of representations for explicitly encoding the temporal patterns in time
series as images. We test our approach on twelve time
series datasets produced from 2D shape, physiological surveillance, industry and other domains. 
Two real spatial-temporal trajectory datasets are also considered for experiments to demonstrate the performance of our approach. We applied deep Convolutional
Neural Networks with a pretraining stage that exploits
local orthogonality by Topographic ICA \cite{ngiam2010tiled} to
``visually'' inspect and classify time series. We report our classification
performance both on GAF and MTF separately, and GAF-MTF which resulted
from combining GAF and MTF representations into single image. By
comparing our results with the current best
hand-crafted representation and classification methods on both time series and trajectory data, we show that
our approach in practice achieves competitive performance with the
state of the art with only cursory exploration of hyperparameters. In
addition to exploring the high level features learned by Tiled CNNs, we
provide an in-depth analysis in terms of the duality between time
series and images. This helps us to more precisely identify
the reasons why our approaches work well.

\section{Motivation}
Learning the (long) temporal correlations that are often embedded in time series remains  a major challenge in time series analysis and modeling. Most real-world data has a temporal component, whether it is measurements of natural (weather, sound) or man-made (stock market, robotics) phenomena. Traditional approaches for modeling and representing time-series data fall into three categories. In time series learning problems, non-data adaptive models, such as Discrete Fourier Transformation (DFT) \cite{agrawal1993efficient}, Discrete Wavelet Transformation (DWT) \cite{chan2003haar}, and Discrete Cosine Transformation (DCT) \cite{korn1997efficiently}, compute the transformation with an algorithm that is invariant with respect to the data to capture the intrinsic temporal correlation with the different basis functions. Meanwhile, researchers explored in the model-based approaches to model time series, such as Auto-Regressive Moving Average models (ARMA) \cite{kalpakis2001distance} and Hidden Markov Models (HMMs) \cite{panuccio2002hidden}, in which the underlying data is assumed to fit a specific type of model to explicitly function the temporal patterns. The estimated parameters can then be used as features for classification or regression. However, more complex, high-dimensional, and noisy real-world time-series data are often difficult to model because the dynamics are either too complex or unknown. Traditional methods, which contain a small number of non-linear operations, might not have the capacity to accurately model such complex systems.

If implicitly learning the complex temporal correlation is difficult, how about reformulating the data to explicitly or even visually encode the temporal dependency, allowing the algorithms to learn more easily? Actually, reformulating the features of time series as visual clues has raised much attention in computer science and physics. The typical examples in speech recognition tasks are that acoustic/speech data input is typically represented by MFCCs or PLPs to explicitly represent the temporal and frequency information. Recently, researchers are trying to build different network structures from time series for visual inspection or designing distance measures. Recurrence Networks were proposed to analyze the structural properties of time series from complex systems \cite{donner2010recurrence,donner2011recurrence}. They build adjacency matrices from the predefined recurrence functions to interpret the time series as complex networks. Silva et al. extended the recurrence plot paradigm for time series classification using compression distance \cite{silva2013time}. Another way to build a weighted adjacency matrix is extracting transition dynamics from the first order Markov matrix \cite{campanharo2011duality}. Although these maps demonstrate distinct topological properties among different time series, it remains unclear how these topological properties relate to the original time series since they have no exact inverse operations. One of our contributions is to propose a set of off-line algorithm to encode the complex correlations in time series into images for visual inspection and classification. The proposed encoding functions have exact/approximate inverse maps, making such transformations more interpretable.  

\section{Encoding Methods}

We first introduce our two frameworks to encode time series data as
images. The first type of image is the Gramian Angular field (GAF), in
which we represent time series in a polar coordinate system instead of
the typical Cartesian coordinates.  In the Gramian matrix, each
element is actually the cosine of the summation of  pairwise temporal values. Inspired by
previous work on the duality between time series and complex networks
\cite{campanharo2011duality}, the main idea of the second framework,
the Markov Transition Field (MTF), is to build the Markov matrix of
quantile bins after discretization and encode the dynamic transition
probability in a quasi-Gramian matrix.

\subsection{Gramian Angular Field}

Given a time series $X = \{x_1, x_2, ..., x_n\}$ of $n$ real-valued
observations, we rescale $X$ so that all values fall in the interval
$[-1,1]$ or $[0,1]$ by:
\begin{eqnarray}
& \tilde{x}_{-1}^i = \frac{(x_i-max(X)+(x_i-min(X))}{max(X)-min(X)}
\label{eqn:rescale-1} \\
\text{or} &\tilde{x}_{0}^i = \frac{x_i-min(X)}{max(X)-min(X)}
\label{eqn:rescale0}
\end{eqnarray}
Thus we can represent the rescaled time series
$\tilde{X}$ in polar coordinates by encoding the value as the
angular cosine and the time stamp as the radius with the equations below: 
\begin{eqnarray}
\left\{\begin{matrix}
\phi=\arccos{(\tilde{x_i})}, -1\leq \tilde{x_i} \leq 1, \tilde{x_i} \in \tilde{X}
\\r= \frac{t_i}{N}, t_i \in \mathbb{N}
\end{matrix}\right.
\label{eqn:polar}
\end{eqnarray}

$t_i$ is the time stamp and $N$ is a constant
factor to regularize the span of the polar coordinate system. This
polar coordinate based representation is a novel way to understand
time series. As time increases, corresponding values warp among
different angular points on the spanning circles, like water
rippling. The encoding map of Eq. \ref{eqn:polar} has two
important properties. First, it is bijective as $\cos(\phi)$ is
monotonic when $\phi \in [0,\pi]$. Given a time series, the proposed
map produces one and only one result in the polar coordinate system
with a unique inverse function. Second, as opposed to Cartesian
coordinates, polar coordinates preserve absolute temporal
relations. In Cartesian coordinates, the area is defined by
$S_{i,j}=\int_{x(i)}^{x_(j)} f(x(t))dx(t)$, we have
$S_{i,i+k}=S_{j,j+k}$ if $f(x(t))$ has the same values on $[i,i+k]$
and $[j,j+k]$. However, in polar coordinates, if the area is defined
as $S'_{i,j}=\int_{\phi(i)}^{\phi(j)}r[\phi(t)]^2d(\phi(t))$, 
then $S'_{i,i+k} \neq S'_{j,j+k}$. That is, the
corresponding area from time stamp $i$ to time stamp $j$ is not only
dependent on the time interval $|i-j|$, but also determined by the
absolute value of $i$ and $j$. 

After transforming the rescaled time series into the polar coordinate
system, we can easily exploit the angular perspective by considering
the trigonometric sum between each pair of points to identify the temporal
correlation in different time intervals. The GAF is defined as
follows: 
\begin{eqnarray}
G   
&=& \begin{bmatrix}
\cos(\phi_1+\phi_1)  & \cdots & \cos(\phi_1+\phi_n) \\
\cos(\phi_2+\phi_1)  & \cdots & \cos(\phi_2+\phi_n) \\
\vdots  & \ddots & \vdots \\
\cos(\phi_n+\phi_1)  & \cdots & \cos(\phi_n+\phi_n) \\
\end{bmatrix} 
\\
&=& \tilde{X}' \cdot \tilde{X} - \sqrt{I-\tilde{X}^2}' \cdot \sqrt{I-\tilde{X}^2} 
\label{eqn:GAF}
\end{eqnarray}

$I$ is the unit row vector $[1,1,...,1]$. After transforming to the polar
coordinate system, we take the data in a time series as a 1-D
metric space. By defining the inner product $<x,y> = x\cdot
y-\sqrt{1-x^2} \cdot \sqrt{1-y^2}$, $G$ is a Gramian matrix:   

\begin{equation}
\begin{bmatrix}
<\tilde{x_1},\tilde{x_1}>  & \cdots & <\tilde{x_1},\tilde{x_n}> \\
<\tilde{x_2},\tilde{x_1}>  & \cdots & <\tilde{x_2},\tilde{x_n}>\\
\vdots  & \ddots & \vdots \\
<\tilde{x_n},\tilde{x_1}>  & \cdots & <\tilde{x_n},\tilde{x_n}> \\
\end{bmatrix} 
\label{eqn:innerproduct}
\end{equation}

\begin{figure}[t]
    \centering
    \includegraphics[scale = 0.33]{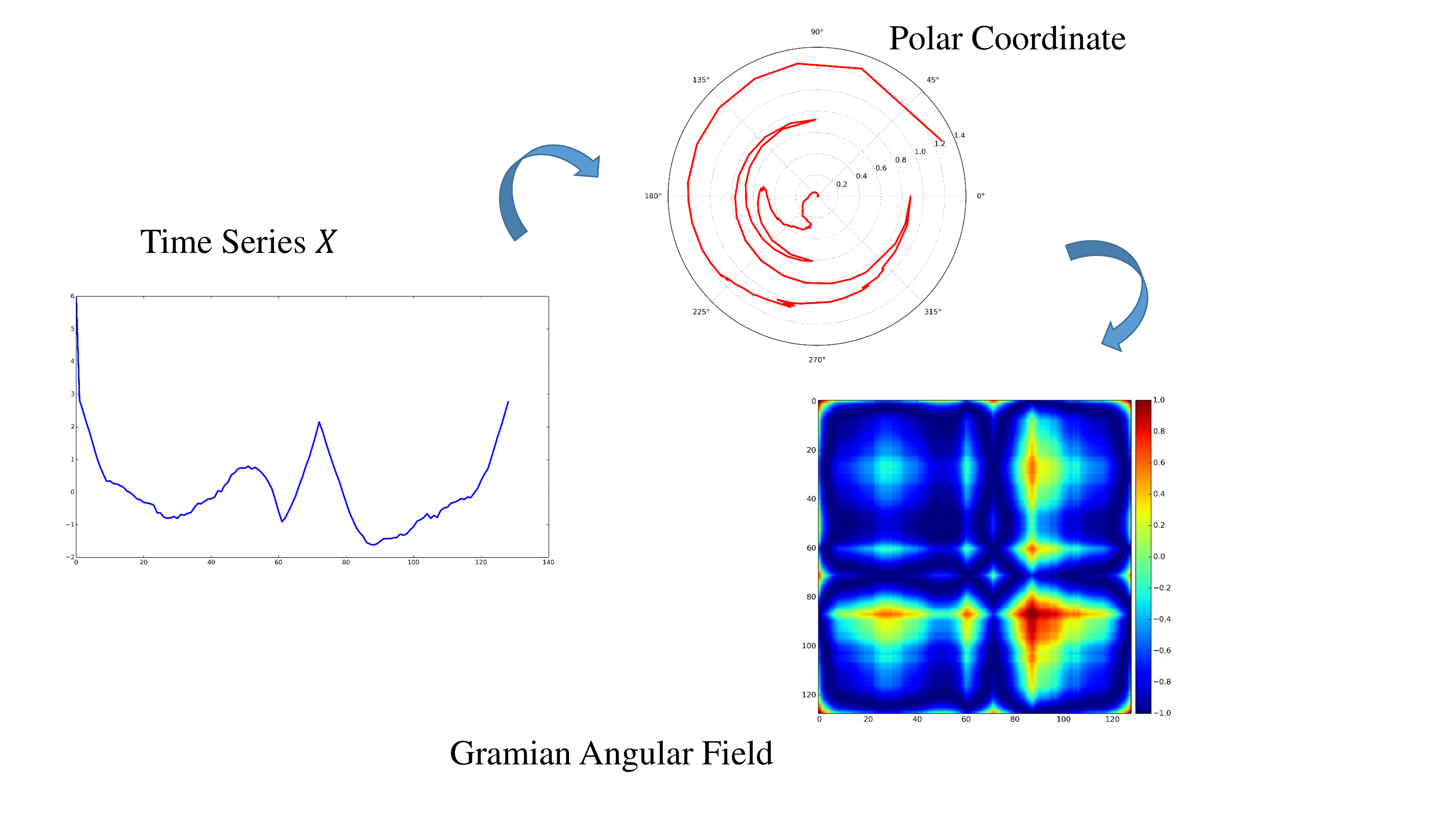}
    \caption{Illustration of the proposed encoding map of the Gramian Angular Field. $X$ is a sequence of typical time series in 'SwedishLeaf' dataset. After $X$ is rescaled by Equation. (\ref{eqn:rescale}) and optionally smoothed by PAA , we transform it to a polar coordinate system by Equation. (\ref{eqn:polar}) and finally calculate its GAF image with Equation. (\ref{eqn:GAF}). In this example, we build GAF without PAA smoothing, so the GAF has a high resolution of $128 \times 128$. } 
    \label{fig:encodingTS2GAF}
\end{figure}

GAF has several advantages. It provides a way to preserve temporal dependency. When time increases, the position moves
from top-left to bottom-right in the Gramian matrix. The GAF contains temporal correlations, as $G_{(i,j||i-j|=k)}$ represents the relative correlation by
superposition of directions with respect to time interval $k$. The
main diagonal $G_{i,i}$ is the special case when $k=0$, which contains
the original value/angular information. With the main diagonal, we
will approximately reconstruct the time series from the high level
features learned by the deep neural network. The GAF images may be large
because the size of the Gramian matrix is $n\times n$ when the length of
the raw time series is $n$. To reduce the size of the GAF images, we apply
Piecewise Aggregate Approximation \cite{keogh2000scaling} to smooth
the time series while keeping the overall trends. The full procedure for
generating the GAF is illustrated in Figure \ref{fig:encodingTS2GAF}.

\begin{figure}[t]
    \centering
    \includegraphics[scale = 0.40]{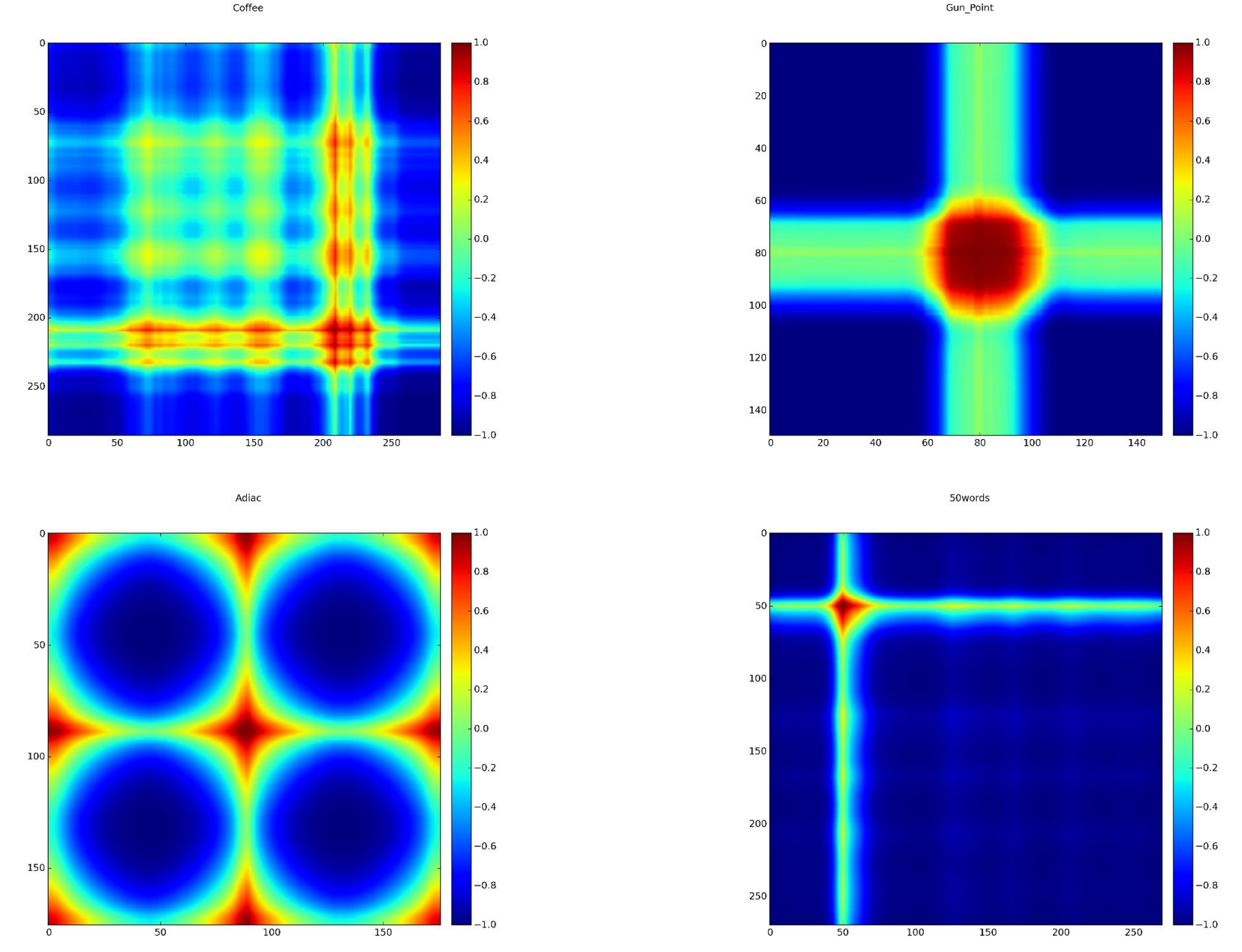}
    \caption{Examples of GAF images on the 'Coffee', 'Gun-Point', 'Adiac' and '50Words' datasets.} 
    \label{fig:GAFdemos}
\end{figure}

Through the polar coordinate system, GAFs actually represent the mutual correlations between each pair of points/phases by the superposition of the nonlinear cosine functions. Different types of time series always have their specific patterns embedded along the time and frequency dimensions. After the feature reformulation process by GAF, most different patterns are enhanced even for visual inspection by humans (Figure \ref{fig:GAFdemos}).    

\subsection{Markov Transition Field}

We propose a framework that is similar to \cite{campanharo2011duality} for
encoding dynamical transition statistics. We develop that idea by
representing the Markov transition probabilities sequentially  to
preserve information in the temporal dimension.  

Given a time series $X$, we identify its $Q$ quantile bins and assign
each $x_i$ to its corresponding bin $q_j$ ($j \in [1,Q]$). Thus we
construct a $Q \times Q$ weighted adjacency matrix $W$ by counting
transitions among quantile bins in the manner of a first-order
Markov chain along the time axis. $w_{i,j}$ is the frequency
with which a point in quantile $q_j$ is followed by a point in quantile
$q_i$. After normalization by $\sum_j{w_{ij}=1}$, $W$ is the Markov
transition matrix:

\begin{equation}
V = 
\begin{bmatrix}
v_{11|P(x_t \in q_1|x_{t-1} \in q_1)}  & \cdots & v_{1Q|P(x_t \in q_1|x_{t-1} \in q_Q)} \\
v_{21|P(x_t \in q_2|x_{t-1} \in q_1)}  & \cdots & v_{2Q|P(x_t \in q_2|x_{t-1} \in q_Q)} \\
\vdots  & \ddots & \vdots \\
v_{Q1|P(x_t \in q_Q|x_{t-1} \in q_1)}  & \cdots & v_{QQ|P(x_t \in q_Q|x_{t-1} \in q_Q)} \\
\end{bmatrix} 
\label{eqn:MTP}
\end{equation} 

It is insensitive to the distribution of $X$ and the temporal dependency on the time steps $t_i$. However, getting rid of the temporal dependency results in too much information loss in
the matrix $W$. To overcome this drawback, we define the Markov Transition
Field (MTF) as follows:
\begin{equation}
M = 
\begin{bmatrix}
v_{ij|x_1 \in q_i,x_1 \in q_j}  & \cdots & v_{ij|x_1 \in q_i,x_n \in q_j} \\
v_{ij|x_2 \in q_i,x_1 \in q_j}  & \cdots & v_{ij|x_2 \in q_i,x_n \in q_j} \\
\vdots  & \ddots & \vdots \\
v_{ij|x_n \in q_i,x_1 \in q_j}  & \cdots & v_{ij|x_n \in q_i,x_n \in q_j} \\
\end{bmatrix} 
\label{eqn:MTF}
\end{equation} 

\begin{figure}[t]
    \centering
    \includegraphics[scale = 0.33]{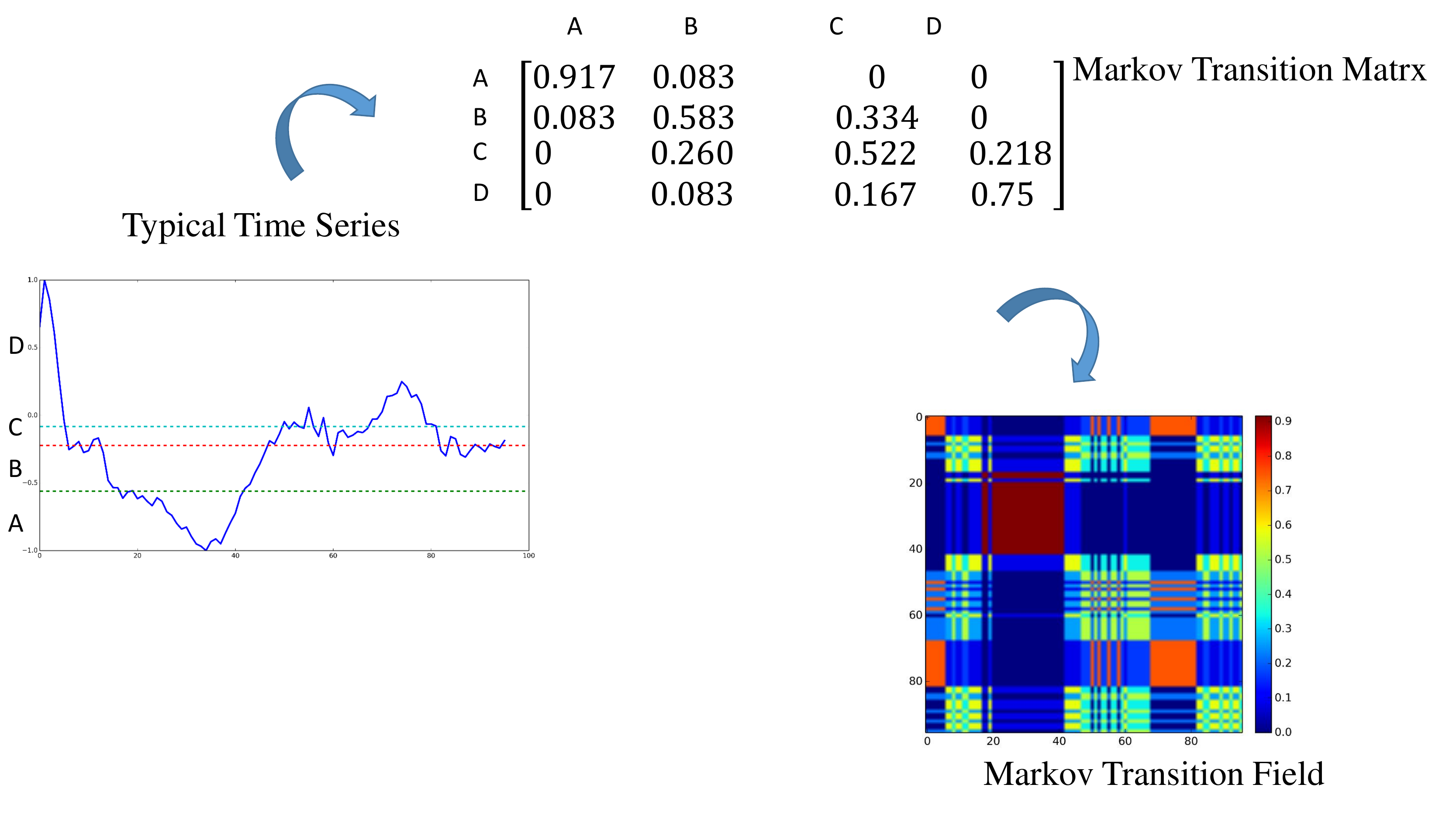}
    \caption{Illustration of the proposed encoding map of a Markov Transition Field. $X$ is a sequence of the  typical time series in the 'ECG' dataset. $X$ is first discretized into $Q$ quantile bins. Then we calculate its Markov Transition Matrix $W$ and finally build its MTF $M$ by Equation. (\ref{eqn:MTF}). We reduce the image size from $96 \times 96$ to $48 \times 48$ by averaging the pixels in each non-overlapping $2 \times 2$ patch.}
    \label{fig:encodingTS2MTF}
\end{figure}

We build a $Q \times Q$ Markov transition matrix $W$ by
dividing the data (magnitude) into $Q$ quantile bins. The quantile
bins that contain the data at time steps $i$ and $j$ (temporal axis)
are $q_{i}$ and $q_{j}$ ($q \in [1,Q]$). $M_{ij}$ in MTF denotes the
transition probability of $q_{i} \rightarrow q_{j}$. That is, we
spread out matrix $W$, which contains the transition probability on the
magnitude axis, into the MTF matrix by considering temporal positions.

By assigning the probability from the quantile at time step $i$ to the
quantile at time step $j$ at each pixel $M_{ij}$, the MTF $M$ actually
encodes multi-step transition probabilities of the time
series. $M_{i,j||i-j|=k}$ denotes the transition probability between
the points with time interval $k$. For example, $M_{ij|j-i=1}$
illustrates the transition process along the time axis with a skip
step. The main diagonal $M_{ii}$, which is a special case when $k=0$
captures the probability from each quantile to itself (the
self-transition probability) at time step $i$. To make the image size
manageable for more efficient computation, we reduce the MTF  by
averaging the pixels in each non-overlapping $m \times m$ patch with
the blurring kernel $\{\frac{1}{m^2}\}_{m \times m}$. That is, we
aggregate the transition probabilities in each subsequence of length
$m$ together. Figure \ref{fig:encodingTS2MTF} shows the procedure to
encode time series to MTF. 

\begin{figure}[t]
    \centering
    \includegraphics[scale = 0.53]{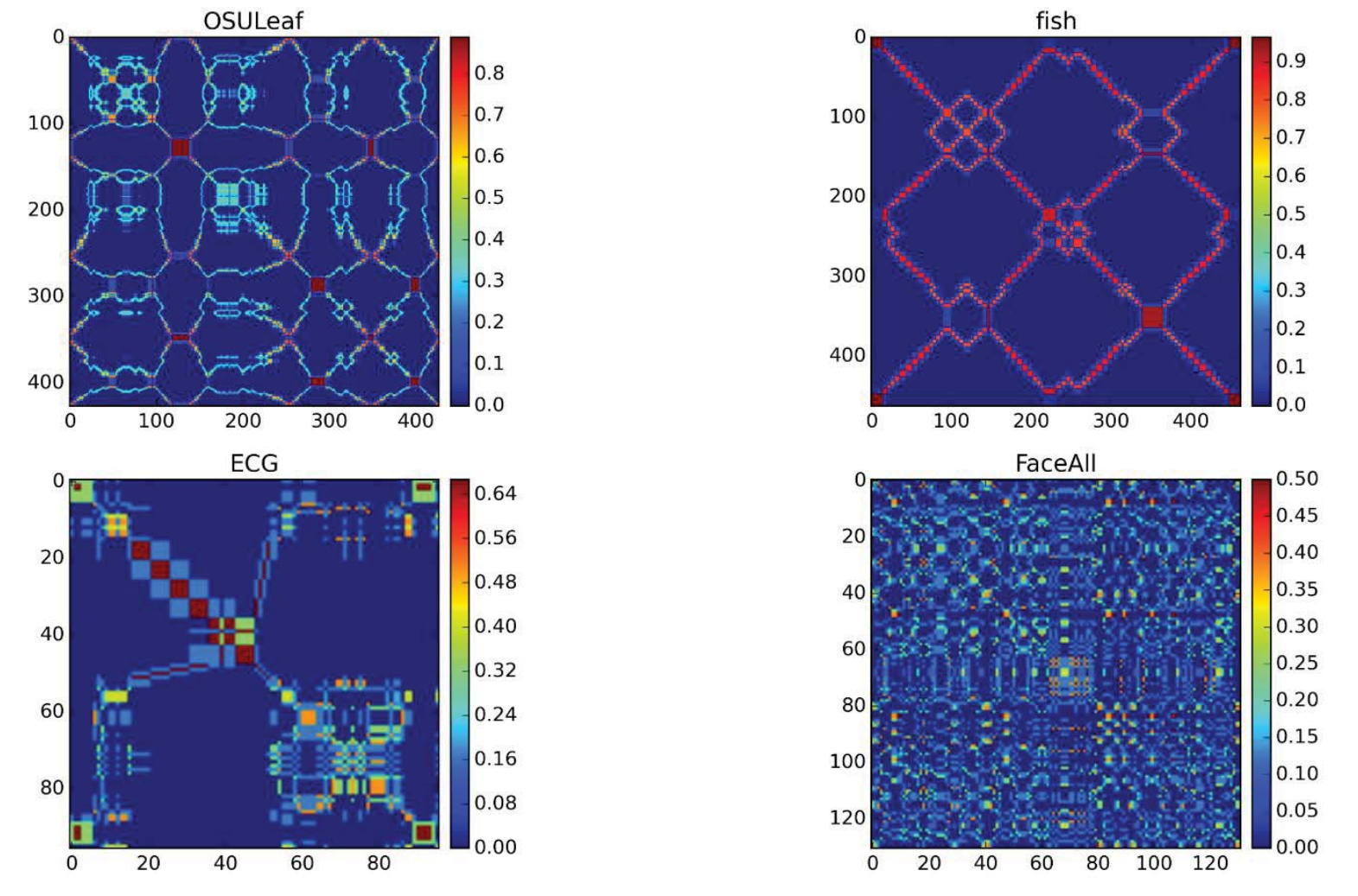}
    \caption{Examples of MTF images on the 'OSUleaf', 'fish', 'ECG' and 'Faceall' datasets.} 
    \label{fig:MTFdemos}
\end{figure}

By scattering the first-order transition probability into the temporally ordered matrix, MTFs encode the transition dynamics between different time lags $k$. We assume that different types of time series have their specific transition dynamics embedded in the temporal and frequency domains. After the feature reformulation process by MTF, most transition dynamics are extracted, which are explicitly obvious for visual inspection (Figure \ref{fig:MTFdemos}).   

\section{Tiled Convolutional Neural Networks} 

Tiled Convolutional Neural Networks \cite{ngiam2010tiled} are a 
variation of  Convolutional Neural Networks. They use tiles and multiple
feature maps to learn invariant features. Tiles are parameterized by a
tile size $K$ to control the distance over which weights are
shared. By producing multiple feature maps, Tiled CNNs learn
overcomplete representations through unsupervised pretraining with
Topographic ICA (TICA).

A typical TICA network is actually a double-stage optimization
procedure with square and square root nonlinearities in each stage,
respectively. In the first stage, the weight matrix $W$ is learned
while the matrix $V$ is hard-coded to represent the topographic
structure of units. More precisely, given a sequence of inputs
$\{x^{h}\}$, the activation of each unit in the second stage is
$f_{i}(x^{(h)};W,V) =
\sqrt{\sum_{k=1}^{p}V_{ik}(\sum_{j=1}^{q}W_{kj}x_j^{(h)})^2}$. TICA
learns the weight matrix $W$ in the second stage by solving:

\begin{equation}
\begin{aligned}
& \underset{W}{\text {minimize}} & & \sum_{h=1}^{n}\sum_{i=1}^{p}f_i(x^{(h)};W,V) \\
& \text{subject to} & & WW^T=I
\end{aligned} 
\label{eq:TICA}
\end{equation}  

$W \in \mathbb{R}^{p \times q}$ and $V \in \mathbb{R}^{p \times p}$ where $p$ is the number of hidden units in a layer and $q$ is
the size of the input. $V$ is a logical matrix ($V_{ij}=1$ or $0$)
that encodes the topographic structure of the hidden units by a contiguous
$3 \times 3$ block. The orthogonality constraint $WW^T=I$ provides
diversity among learned features. 

\begin{algorithm}[t]
\begin{algorithmic}
\Require {$\{x^{(t)}\}^T_{t=1}, v, s, W, V$ as input}
\Ensure {$W$ as output}
\Repeat \State $f^{old} = \sum_{t=1}^{T} \sum_{i=1}^{m} \sqrt{\sum_{k=1}^{m}V_{ik}(\sum_{j=1}^{n}W_{kj}x^{(t)}_j)^2}$,
$g = \frac{f^{old}}{\partial W}$, $f^{new} = + \infty$, $\alpha = 1$ 
\While {$f^{new} > f^{old}$} \\
\hskip\algorithmicindent \hskip\algorithmicindent $W^{new} = W - \alpha g $ \\
\hskip\algorithmicindent \hskip\algorithmicindent $W^{new} = Localize(W^{new}, s) $ \\
\hskip\algorithmicindent \hskip\algorithmicindent $W^{new} = tieWeights(W^{new}, k) $ \\
\hskip\algorithmicindent \hskip\algorithmicindent $W^{new} = orthogonalizeLocalRF(W^{new}) $ \\
\hskip\algorithmicindent \hskip\algorithmicindent $W^{new} = tieWeights(W^{new}, k) $ \\
\hskip\algorithmicindent \hskip\algorithmicindent $f^{new} = \sum_{t=1}^{T} \sum_{i=1}^{m} \sqrt{\sum_{k=1}^{m}V_{ik}(\sum_{j=1}^{n}W_{kj}x^{(t)}_j)^2}$ \\
\hskip\algorithmicindent \hskip\algorithmicindent $ \alpha = 0.5 \alpha$

\EndWhile \\
\hskip\algorithmicindent $W = W^{new}$
\Until {convergence}
\end{algorithmic}
\caption{Unsupervised pretraining with TICA \cite{ngiam2010tiled}}
\label{alg:TICA}
\end{algorithm}

The pretraining algorithm (Algorithm. \ref{alg:TICA}) is based on gradient descent on the TICA objective function in Equation. \ref{eq:TICA}. The inner loop is a simple implementation of backtracking linesearch. The $orthogonalize\_localRF(W^new)$ function only orthogonalizes the weights that have completely overlapping receptive fields.  Weight-tying is applied by averaging each set of tied weights. The algorithm is trained by batch projected gradient descent. Other unsupervised feature learning algorithms such as RBMs and autoencoders \cite{bengio2007greedy} require more parameter tuning, especially during optimization. However, pretraining with TICA usually requires little tuning of optimization parameters, because the tractable objective function of TICA allows to monitor convergence easily. 
 
Neither GAF nor MTF images are natural images; they have no natural concepts
such as ``edges'' and ``angles''. Thus, we propose to exploit the
benefits of unsupervised pretraining with TICA to learn many diverse
features with local orthogonality. In
\cite{ngiam2010tiled}, the authors empirically demonstrate that tiled CNNs
perform well with limited labeled data because the partial weight
tying requires fewer parameters and reduces the need for a large
amount of labeled data. Our data from the UCR Time Series Repository
\cite{keogh2011ucr} tends to have few instances (e.g., the ``yoga''
dataset has 300 labeled instance in the training set and 3000
unlabeled instance in the test set), so tiled CNNs are suitable for our
learning task. Moreover, Tiled CNNs achieve good performance on large datasets (such as NORB and CIFAR). 

\begin{figure}[t]
    \centering
    \includegraphics[scale = 0.33]{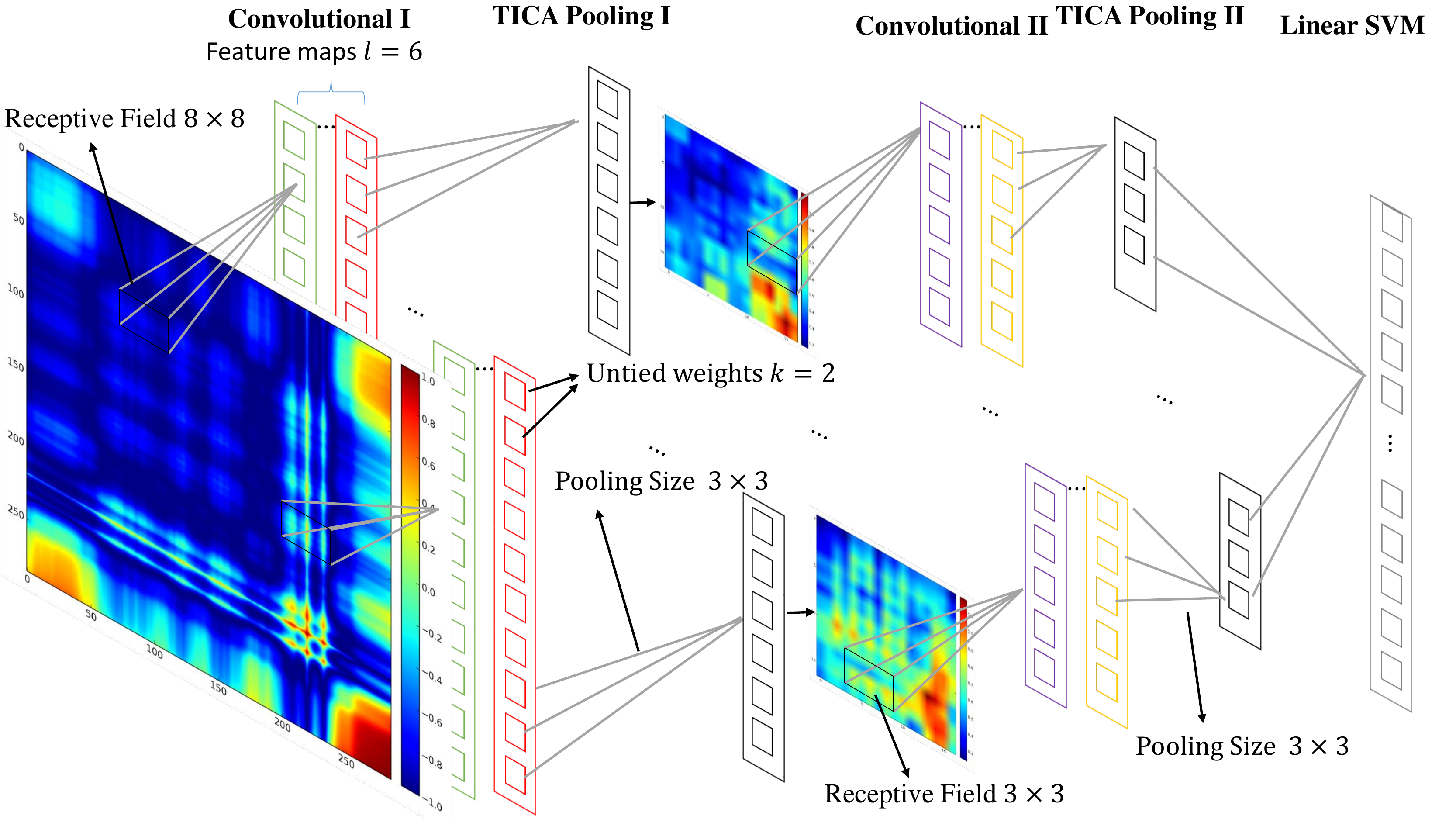}
    \caption{Structure of the tiled convolutional neural network. 	We
      fix the size of receptive fields to $8 \times 8$ in the first
      convolutional layer and $3 \times 3$ in the second convolutional
      layer. Each TICA pooling layer pools over a block of $3 \times
      3$ input units in the previous layer without wraparound at the boarders to optimize for sparsity of the pooling units. The number of pooling units in each map is exactly the same as the number of input units. The last layer is a linear SVM for classification. We construct this network by stacking two Tiled CNNs, each with 6 maps ($l=6$) and tiling size $k=2$.}
    \label{fig:TCNN_structure}
\end{figure}

Typically, tiled CNNs are trained with two hyperparameters, the tiling
size $k$ and the number of feature maps $l$. In our experiments, we
directly fixed the network structures without tuning these
hyperparameters in loops. Our experimental settings follow 
the default deep network structures and parameters in \cite{ngiam2010tiled}. Tiled CNNs with such configurations are reported to achieve the best performance on the NORB image classification benchmark. Although tuning the parameters
will surely enhance performance, doing so may cloud our understanding
of the power of the representation. Another consideration is
computational efficiency. All of the experiments on the 12
datasets could be done in one day on a laptop with an Intel i7-3630QM
CPU and 8GB of memory (our experimental platform). Thus, the results
in this paper are a preliminary lower bound on the potential best
performance. Thoroughly exploring network structures and
parameters will be addressed in future work. The structure and
parameters of the tiled CNN used in this paper are illustrated in
Figure \ref{fig:TCNN_structure}.

\section{Experiments on Time Series Data}
\label{sec:UCR}

\begin{table}[t]
  \centering
  \caption{Summary statistics of 12 standard datasets}
    \begin{tabular}{rrrrr}
    \toprule
    DATASET & CLASS & TRAIN & TEST & LENGTH \\
    \midrule
    50words & 50 & 450 & 455 & 270 \\
    Adiac & 37 & 390 & 391 & 176 \\
    Beef & 5 & 30 & 30 & 470 \\
    Coffee & 2 & 28 & 28 & 286 \\
    ECG200 & 2 & 100 & 100 & 96 \\
    FaceAll & 14 & 560 & 1,690 & 131 \\
    Lightning2 & 2 & 60 & 61 & 637 \\
    Lightning7 & 7 & 70 & 73 & 319 \\
    OliveOil & 4 & 30 & 30 & 570 \\
    OSULeaf & 6 & 200 & 242 & 427 \\
    SwedishLeaf & 15 & 500 & 625 & 128 \\
    Yoga & 2 & 300 & 3,000 & 426 \\
    \bottomrule
    \end{tabular}%
  \label{tab:UCRstat}%
\end{table}%


We apply Tiled CNNs to classify using GAF and MTF representation on
twelve tough datasets, on which the classification error rate is above
$0.1$ with the state-of-the-art SAX-BoP approach
\cite{lin2012rotation,oates2012exploiting}. More detailed statistics
are summarized in Table \ref{tab:UCRstat}. The datasets are
pre-split into training and testing sets for experimental
comparisons. For each dataset, the table gives its name, the number of
classes, the number of training and test instances, and the length of
the individual time series.  

\subsection{Experiment Settings}

In our experiments, the size of the GAF image is regulated by the the
number of PAA bins $S_{GAF}$. Given a time series $X$ of size $n$, we
divide the time series into $S_{GAF}$ adjacent, non-overlapping
windows along the time axis and extract the means of each bin. This
enables us to construct the smaller GAF matrix $G_{S_{GAF} \times
  S_{GAF}}$. MTF requires the time series to be discretized into $Q$
quantile bins to calculate the $Q \times Q$ Markov transition matrix,
from which we construct the raw MTF image $M_{n \times n}$
afterwards. Before classification, we shrink the MTF image size to
$S_{MTF} \times S_{MTF}$ by the blurring kernel $\{\frac{1}{m^2}\}_{m
\times  m}$ where $m=\lceil \frac{n}{S_{MTF}}\rceil$. The Tiled CNN
is trained with image size $\{S_{GAF},S_{MTF}\} \in \{16, 24, 32, 40, 48\}$ and quantile size $Q \in \{8, 16, 32, 64\}$. At the last layer
of the Tiled CNN, we use a linear soft margin SVM
\cite{fan2008liblinear} and select $C$ by 5-fold cross validation
over $\{10^{-4}, 10^{-3}, \ldots, 10^4\}$ on the training set. 

For each input of image size $S_{GAF}$ or $S_{MTF}$ and quantile size
$Q$, we pretrain the Tiled CNN with the full unlabeled dataset (both training and test set with no labels) to learn
the initial weights $W$ through TICA.Then we train the SVM at the
last layer by selecting the penalty factor $C$ with cross
validation. Finally, we classify the test set using the optimal
hyperparameters $\{S, Q, C\}$ with the lowest error rate on the
training set. If two or more models tie, we prefer the larger $S$
and $Q$ because larger $S$ helps preserve more information through
the PAA procedure and larger $Q$ encodes the dynamic transition
statistics with more detail. Our model selection approach provides
generalization without being overly expensive computationally.

\subsection{Results and Discussion}

\begin{table}[t]
\small
  \centering
  \caption{Tiled CNN error rate on training set and test set}
    \begin{tabular}{rrrrr}
    \toprule
    \multicolumn{1}{c}{DATASET} & \multicolumn{2}{c}{GAF} & \multicolumn{2}{c}{MTF} \\
    \midrule
    \multicolumn{1}{c}{} & TRAIN  & TEST  & TRAIN  & TEST  \\
    50words & 0.338 & 0.310 & 0.442 & 0.426 \\
    adiac & 0.321 & 0.284 & 0.638 & 0.665 \\
    beef & 0.633 & 0.4 & 0.533 & 0.233 \\
    coffee & 0 & 0 & 0 & 0 \\
    ECG200 & 0.16 & 0.11 & 0.15 & 0.21 \\
    faceall & 0.121 & 0.244 & 0.102 & 0.259 \\
    lighting2 & 0.2 & 0.18 & 0.167 & 0.361 \\
    lighting7 & 0.329 & 0.397 & 0.386 & 0.411 \\
    oliveoil & 0.2 & 0.2 & 0.033 & 0.3 \\
    OSULeaf & 0.415 & 0.463 & 0.43 & 0.483 \\
    SwedishLeaf & 0.134 & 0.104 & 0.206 & 0.176 \\
    yoga & 0.183 & 0.177 & 0.193 & 0.243 \\
    \bottomrule
    \end{tabular}%
  \label{tab:trainteststat}%
\end{table}%

We use Tiled CNNs to classify GAF and MTF representations separately
on the 12 datasets. The training and test error rates are shown in
Table \ref{tab:trainteststat}. Generally, our approach is not prone to
overfitting as seen by the relatively small difference between
training and test set errors.  One exception is the 'Olive Oil' dataset
with the MTF approach where the test error is significantly higher.

In addition to the slight risk of potential overfitting, MTF has generally
higher error rates than GAF. This is most likely because of
uncertainty in the inverse image of MTF. Note that the encoding
function from time series to GAF and MTF are both surjection. The map
functions of GAF and MTF will each produce only one image with
fixed $S$ and $Q$ for each given time series $X$. Because they are both surjective mapping functions, the inverse
image of the map is not fixed. As shown in a later
section, we can approximately reconstruct the raw time series from
GAF, but it is very hard to even roughly recover the signal from
MTF. GAF has smaller uncertainty in the inverse image of its mapping
function because randomness only comes from the ambiguity of
$\cos(\phi)$ when $\phi \in [0,2\pi]$. MTF, on the other hand, has a
much larger inverse image space, which results in large variation when
we try to recover the signals. Although MTF encodes the transition
dynamics, which are important features of time series, such features
seem not to be sufficient for recognition/classification tasks.

Note that at each pixel, $G_{ij}$, denotes the superstition of the
directions at $t_i$ and $t_j$, $M_{ij}$ is the transition probability
from the quantile at $t_i$ to the quantile at $t_j$. GAF encodes 
static information while MTF depicts information about dynamics. From this
point of view, we consider them as two ``orthogonal'' channels, like
different colors in the RGB image space.  Thus, we can combine GAF and MTF
images of the same size (i.e. $S_{GAF}=S_{MTF}$) to construct a
double-channel image (GAF-MTF). Since GAF-MTF combines both the static
and dynamic statistics embedded in raw time series, we posit that it
will improve classification performance. In the next
experiment, we pretrained and fine-tuned the Tiled CNN on the compound
GAF-MTF images. Then, we report the classification error rate on test
sets. 

\begin{table}[t]
\small
  \centering
    \caption{Summary of the error rates from 6 recently published best results and our approach. The symbols $\ast$,  $\triangleleft$, $\dagger$ and $\bullet$ represent datasets generated from figure shapes (2D), physiological surveillance, industry and all remaining temporal signals, respectively.}
    \begin{tabular}{rrrrrrrr}
    \toprule
    Dataset  & 1NN- & 1NN- & Fast  & SAX- & SAX- & RPCD & GAF- \\
      & Euclidean & DTW & shapelet & BoP & VSM &   & MTF \\
     \midrule
    50words $\bullet$ & 0.369 & \textbf{0.242} & 0.4429 & 0.466 & N/A & 0.226 & 0.284 \\
    Adiac $\ast$ & 0.389 & 0.391 & 0.514 & 0.432 & 0.381 & 0.384 & \textbf{0.307} \\
    Beef $\bullet$ & 0.467 & 0.467 & 0.447 & 0.433 & 0.33 & 0.367 & \textbf{0.3} \\
    Coffee $\bullet$ & 0.25 & 0.18 & 0.067 & 0.036 & \textbf{0} & \textbf{0} & \textbf{0} \\
    ECG200 $\triangleleft$ & 0.12 & 0.23 & 0.227 & 0.14 & 0.14 & 0.14 & \textbf{0.08} \\
    FaceAll  $\ast$ & 0.286 & 0.192 & 0.402 & 0.219 & 0.207 & \textbf{0.191} & 0.223 \\
    Lightning2 $\dagger$ & 0.246 & \textbf{0.131} & 0.295 & 0.164 & 0.196 & 0.246 & 0.18 \\
    Lightning7 $\dagger$ & 0.425 & \textbf{0.274} & 0.403 & 0.466 & 0.301 & 0.356 & 0.397 \\
    OliveOil $\bullet$ & 0.133 & 0.133 & 0.213 & 0.133 & \textbf{0.1} & 0.167 & 0.167 \\
    OSULeaf  $\ast$ & 0.483 & 0.409 & 0.359 & 0.236 & \textbf{0.107} & 0.355 & 0.446 \\
    SwedishLeaf  $\ast$ & 0.213 & 0.21 & 0.27 & 0.198 & 0.251 & 0.098 & \textbf{0.093} \\
    Yoga  $\ast$ & 0.17 & 0.164 & 0.249 & 0.17 & 0.164 & \textbf{0.134} & 0.16 \\
    \textit{WINS}\# & \textit{0} & \textit{3} & \textit{0} & \textit{0} & \textit{3} & \textit{3} & \textit{\textbf{5}} \\
    \bottomrule
    \end{tabular}%
  \label{tab:fullstatistics}%
\end{table}%

Table \ref{tab:fullstatistics} compares the classification error rate
of our approach with previously published results of five
competing methods: two state-of-the-art 1NN classifiers based on
Euclidean distance and DTW, the recently proposed Fast-Shapelets based
classifier \cite{rakthanmanon2013fast}, the classifier based on
Bag-of-Patterns (BoP) \cite{lin2012rotation,oates2012exploiting} and
the most recent SAX-VSM approach \cite{senin2013sax}. Our approach
outperforms 1NN-Euclidean, fast-shapelets, and BoP, and is competitive
with 1NN-DTW and SAX-VSM. 

In addition, by comparing the results between Table
\ref{tab:fullstatistics} and Table \ref{tab:trainteststat}, we
verified our assumption that combined GAF-MTF images have better
expressive power than the single GAF or MTF alone for classification. GAF-MTF images
achieves the lower test error rate on ten datasets out of twelve (except
for the 'Adiac' and 'Beef' dataset ). On  the 'Olive Oil' dataset, the
training error rate is $6.67\%$ and the test error rate is
$16.67\%$. This demonstrates that the integration of both types of images 
into one compound image decreases the risk of overfitting as well as
enhancing the overall classification accuracy. 
Thus, the
intrinsic "orthogonality" between GAF and MTF on the same time series
helps improve the classification performance with more comprehensive features. The
multi-channel encoding approach is a scalable framework. The
combination of multiple orthogonal channels into one images potentially
improve the classification results, decreasing the risk of overfitting by a generalized ensemble framework. Meanwhile, hand-crafted feature integration potentially helps learn different
informative features through deep learning architectures.

\subsection{Analysis of Learned Features}

\begin{figure}[t!]
    \centering
    \includegraphics[scale = 0.35]{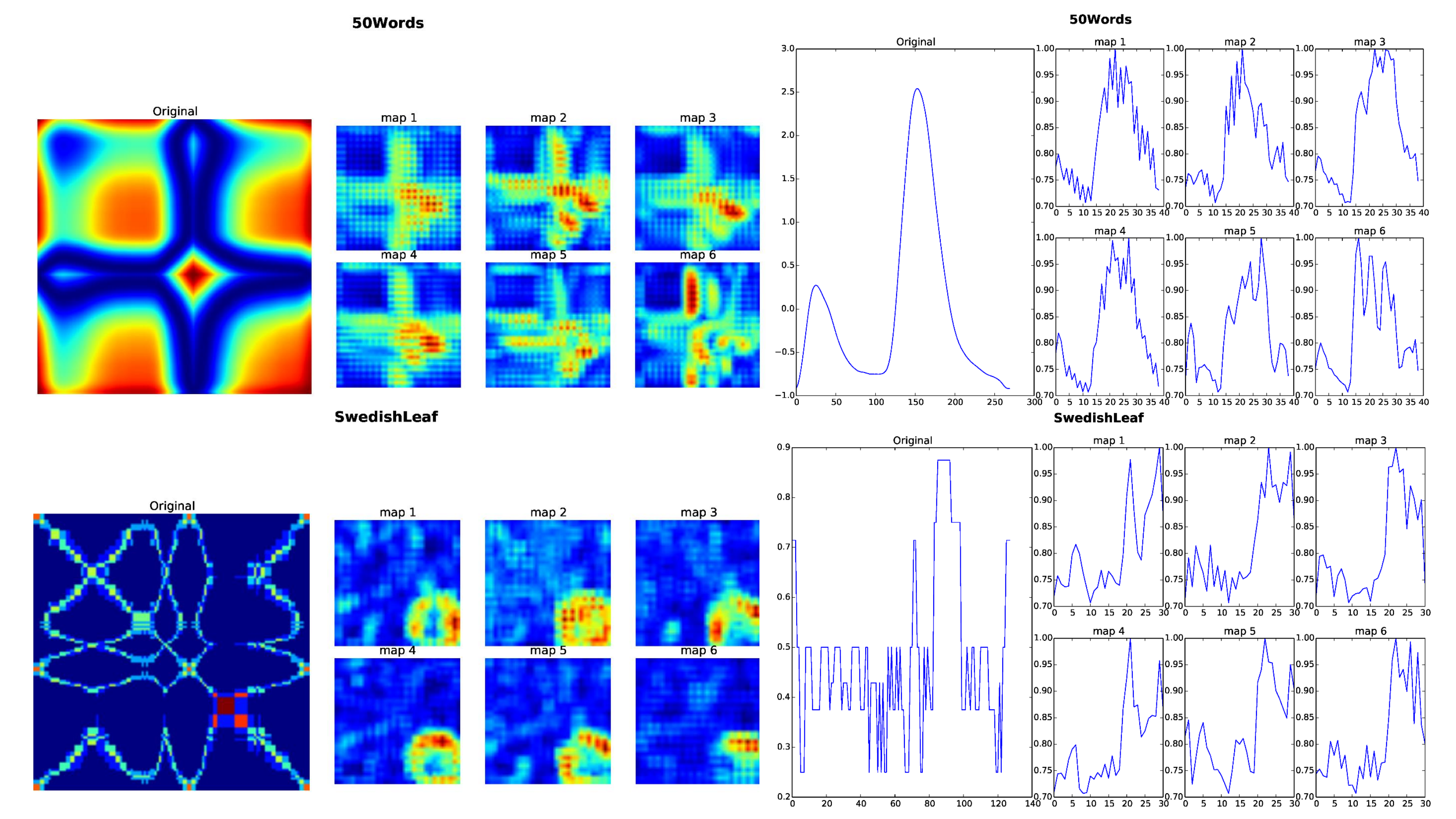}
    \caption{(a) Original GAF and its six learned feature maps before the SVM layer in Tiled CNN (top left), and (b) raw time series and approximate reconstructions based on the main diagonal of six feature maps (top right) on '50Words' dataset; (c) Original MTF and its six learned feature maps before the SVM layer in Tiled CNN (bottom left), and (d) curve of self-transition probability along time axis (main diagonal of MTF) and approximate reconstructions based on the main diagonal of six feature maps (bottom right) on "SwedishLeaf" dataset.}
    \label{fig:feature_reconstruction}
\end{figure}

In contrast to the cases in which the CNN is applied in natural image
recognition tasks, neither GAF nor MTF have natural interpretations of
visual concepts (e.g., ''edges'' or ``angles'').  In this section, we
analyze the features and weights learned through the Tiled CNNs to explain
why our approach works.  

As mentioned earlier, the mapping function from time series to GAF is
surjective and the uncertainty in its inverse image comes from the
ambiguity of  $\cos(\phi)$  when $\phi \in [0,2\pi]$. The main
diagonal of GAF, i.e. $\{G_{ii}\} =\{\cos(2\phi_{i})\}$ allows us to
approximately reconstruct the original time series, ignoring the signs,
by  
\begin{equation}
\cos(\phi) = \sqrt{\frac{\cos(2\phi)+1}{2}}
\end{equation}   

MTF has much larger uncertainty in its inverse image, making it hard
to reconstruct the raw data from MTF alone. However, the diagonal
$\{M_{ij||i-j|=k}\}$ represents the transition probability among the
quantiles in temporal order considering the time interval $k$. We
construct the self-transition probability along the time axis from the
main diagonal of MTF like we do for GAF. Although such reconstructions
less accurately capture the morphology of the raw time series, they
provide another perspective of how Tiled CNNs capture the transition
dynamics embedded in MTF.

Figure \ref{fig:feature_reconstruction} illustrates the reconstruction
results from six feature maps learned before the last SVM layer on the GAF
and MTF. The Tiled CNN extracts the color patch, which is essentially an adaptive 
moving average that enhances several receptive fields within the nonlinear
units by different trained weights. It is not a simple moving average but the synthetic
integration by considering the 2D temporal dependencies among
different time intervals, which is a benefit from the Gramian 
matrix structure that helps preserve the temporal information. By observing the rough orthogonal reconstruction from each layer of the feature maps, we can clearly observe that the CNNs can extract the multi-frequency dependencies through the convolution and pooling architecture on the GAF and MTF images.  Different feature maps preserve the overall trend while addressing more details in different subphases. As shown in
Figures \ref{fig:feature_reconstruction}(b) and
\ref{fig:feature_reconstruction}(d), the high-leveled feature maps
learned by the Tiled CNN are equivalent to a multi-frequency approximator
of the original curve. 

\begin{figure}[t]
    \centering
    \includegraphics[scale = 0.35]{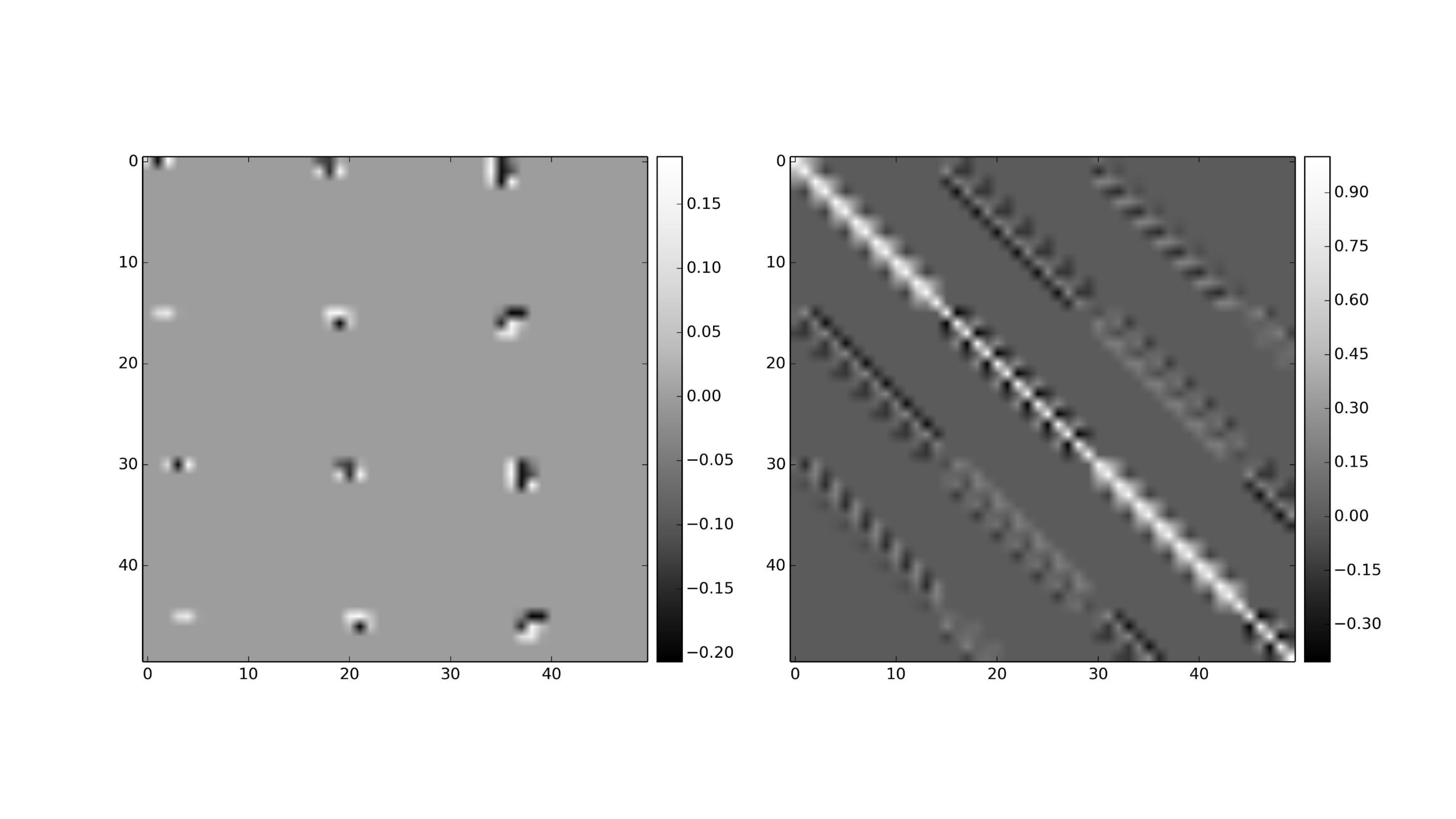}
    \caption{learned sparse weights $W$ for the last SVM layer in Tiled CNN (left) and its  orthogonality constraint by $WW^T=I$ (right).}
    \label{fig:TCNNweights}
\end{figure}

Figure \ref{fig:TCNNweights} demonstrates the
learned sparse weight matrix $W$ with the constraint $WW^T=I$, which
makes effective use of local orthogonality. The TICA pretraining
provides the built-in advantage that the function w.r.t the parameter
space is not likely to be ill-conditioned as $WW^T=1$. As shown in Figure \ref{fig:TCNNweights} (right), the weight matrix $W$ is quasi-orthogonal and approaching 0 without very large
magnitude. This implies that the condition number of $W$ approaches 1 and helps the system to be well-conditioned.

\section{Experiments on Trajectory Data}
We have demonstrated the effectiveness of GAF and MTF the benchmark time series datasets as diverse as shape, physiological surveillance and industry from the UCR time series repository. In this section we describe an application of our approaches to classify  spatial-temporal trajectory data. The trajectory data is complex because patterns of movement are often driven by unperceived goals and constrained by an unknown environment. 

\begin{figure}[t]
    \centering
    \includegraphics[scale = 0.4]{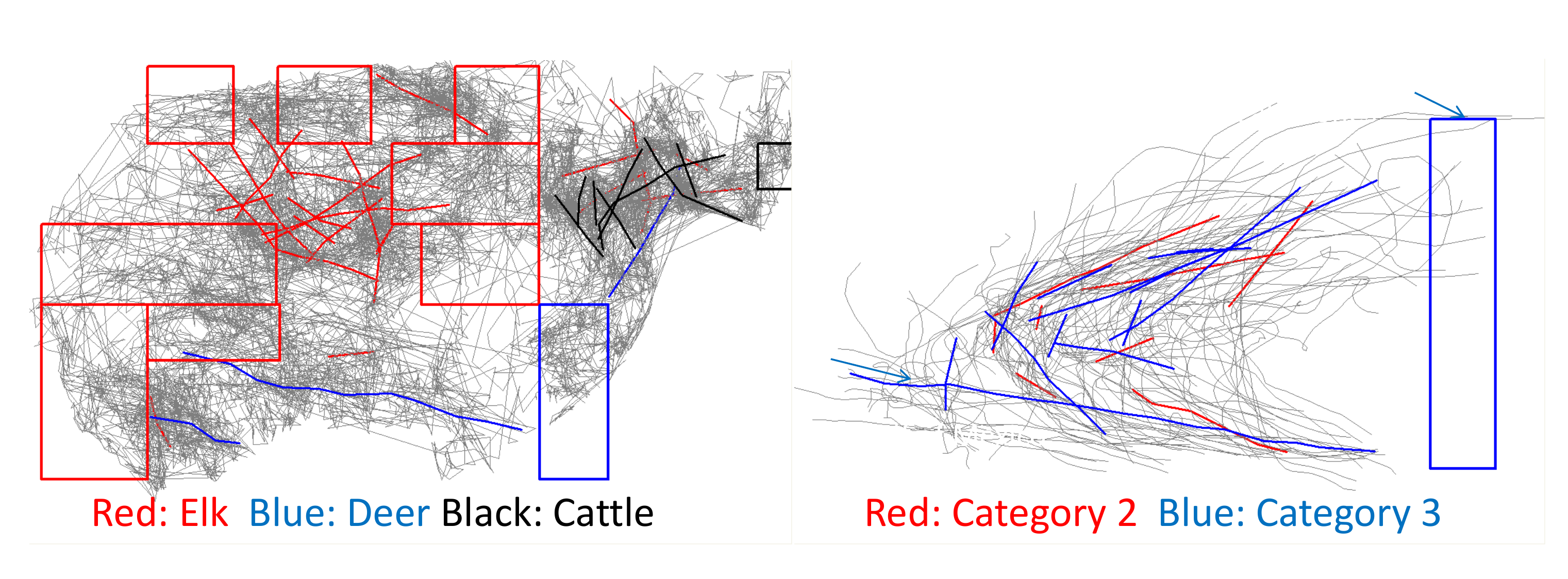}
    \caption{Overview of the trajectory and the RB-TB features \cite{lee2008traclass} learnt in (a). Animal tracking data (left) and (b). Hurricane data (right)}
    \label{fig:trajdata}
\end{figure}

To compare our results with other benchmark approaches including the seminal work from \cite{lee2008traclass}, we run experiments on two benchmark datasets, the animal movement dataset (Animal) and the hurricane track dataset (Hurricane) (Figure \ref{fig:trajdata}). Both datasets have trajectories of unequal length. For the "Animal" dataset, the x and y coordinates are extracted from animal movements observed in June 1995. It is divided into three classes by species: elk, deer, and cattle, as shown in Figure 15. The numbers of trajectories (points) are 38 (7117), 30 (4333), and 34 (3540), respectively. In the "Hurricane" dataset, the latitude and longitude are extracted from Atlantic hurricanes for the years 1950 through 2006. The Saffir-Simpson scale classifies hurricanes into categories 1-5 by intensity. A high category number indicates high intensity. Categories 2 and 3 are chosen for two classes. The numbers of trajectories (points) are 61 (2459) and 72 (3126), respectively. Both datasets are pre-split into two parts for training (80\%) and testing (20\%). Figure \ref{fig:trajdata} shows the overview of the trajectory data. Table \ref{tab:traj} provides the classes, training size, testing size, minimum length and maximum length of the trajectory data.

\begin{table}[b]
  \centering
  \caption{Summary statistics of two trajectory datasets.}
    \begin{tabular}{rrrrrr}
    \toprule
      Dataset & Classes & Training  & Testing  & Min  & Max  \\    
      &   & Size & Size & Length & Length \\
    \midrule
    Animal Tracking & 3 & 80 & 18 & 10 & 291 \\
    Hurricane & 2 & 112 & 21 & 11 & 108 \\
    \bottomrule
    \end{tabular}%
  \label{tab:traj}%
\end{table}%

\subsection{Hilbert Space Filling Curves}
\begin{figure}[t]
    \centering
    \includegraphics[scale = 0.39]{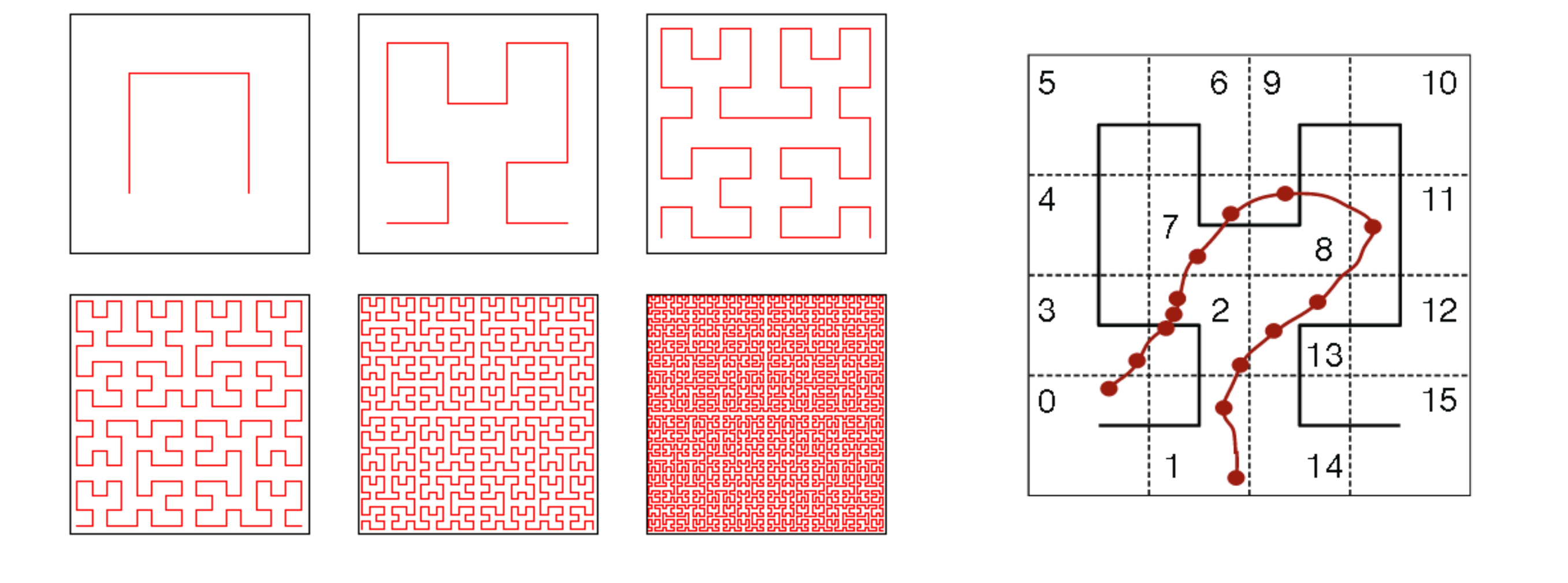}
    \caption{(a). Hilbert space filling curve of order \{1,2,3,4,5,6\} in 2-dimensional space (left) (b). An example of the transformation from 2-dimensional trajectory to 1-dimensional time series using HSCF of order 2 (right).}
    \label{fig:HSFC}
\end{figure}

Spatial-temporal trajectory data is commonly multi-dimensional. We use Hilbert Space Filling Curves (SFC) to transform the trajectory into time series while preserving the spatial-temporal information.

Space filling has been studied by the mathematicians since the late 19th century when the first graphical representation was proposed by David Hilbert in 1891 \cite{hilbert1891ueber}. Space filling curves provide a linear mapping from the multi-dimensional space to the 1-dimensional space. This mapping can be thought of as dividing D-dimensional space into D-dimensional hypercubes with a line passing through each hypercube. Recently, filling curve based approaches have shown to be able to preserve locality between objects in the multidimensional space in the linear space, and thus have been applied to different tasks like clustering \cite{moon2001analysis}, high dimensional outlier detection \cite{angiulli2005outlier}, and trajectory query \cite{ding2008efficient} and classification \cite{gandi2015generative}. Figure \ref{fig:HSFC} (a) shows SFC examples of order \{1,2,3,4,5,6\}. 

Basically, the SFC of order 1 divides the square into 4 area. For the Hilbert curve with order 2, each sub-area of the curve with order 1 is further divided into 4 sub-areas. This process goes on as the order of the SFC increases. It is clear that the number of sub-areas in 2 dimensional SFC is $4^{order}$. To convert 2-dimensional data points to 1-dimensional points, each sub-area is integer numbered from 0 to $4^{order}-1$ starting from the lower left corner as 0 to the lower right corner. All other sub-areas are numbered in order of occurrence of the corresponding vertex as shown in Figure \ref{fig:HSFC} (b) when order = 2. It also shows the example transformation process from a 2D trajectory to a sequence of scalars (time series). The final time series generated after SFC transformation is T = [0, 3, 2, 2, 2, 7, 7, 8, 11, 13, 13, 2, 1, 1].

We map the trajectory points by the visiting order of the SFC embedded in the trajectory manifold space to the index sequence by the recorded times. The produced time series can be used for classification using our algorithm. This adds another hyperparameter called the SFC order, which decides the granularity of the space filling curve. 

\subsection{Experiment Settings}

The parameter settings are the same as the previous experiments on UCR datasets (Section \ref{sec:UCR}). The optimal SFC order is selected together with other parameters through 5-fold cross validation from \{3,4,5,6,7,8,9,10\}. 

Note that both trajectory datasets have quite small sample size with varying length. When the trajectory length (as well as the time series length produced by SFC) is smaller than image size $S$, we uniformly duplicate each point in the time series in temporal order to stretch the sequence to length $S$. If the difference between the length of a time series and $S$ is smaller than the original time series length, the interpolation strategy changes to random duplication instead of following the temporal order.     

\subsection{Results and Discussion}
\begin{figure}[t]
    \centering
    \includegraphics[scale = 0.39]{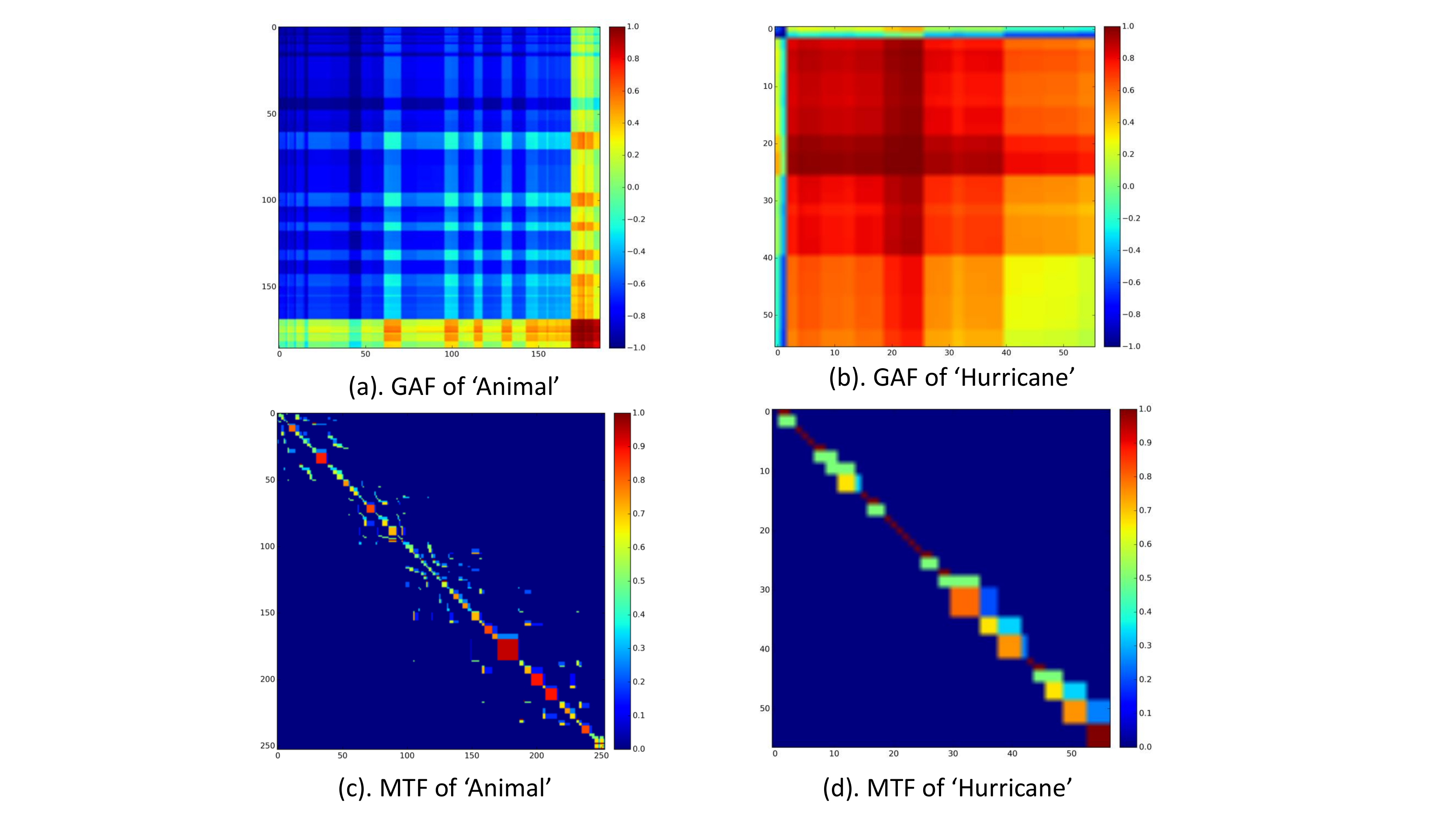}
    \caption{Examples of GAF and MTF images generated from the time series on 'Animal' and 'Hurricane' datasets. The time series is produced using SFC from raw 2D trajectory.}
    \label{fig:trajGAFMTF}
\end{figure}  

Both 'Animal' and 'Hurricane' datasets have been used in previous research \cite{lee2008traclass,gandi2015generative} to achieve state-of-the-art classification accuracy. Traclass give two algorithms, trajectory-based (TB-only) and region-based + trajectory-based  (RB-TB) approaches based on features used for classification on these datastes. They carefully designed a hierarchy of features by partitioning trajectories and exploring  two types of clustering. In \cite{gandi2015generative}, the author used SFC transformation to linearly map the trajectory data to time series and classified the sequences based on symbolic discretization with the multiple normal distribution assumption.  

After transforming the 2D trajectory data to time series using SFC, we generate the corresponding GAF and MTF images as shown in Figure \ref{fig:trajGAFMTF}. However, we found significant overfitting with CNNs even using 5-fold cross validation. This is probably because both the sample size and the time series length of the trajectory datasets are too small to avoid overfitting in neural networks. Previous work has discussed overfitting during cross validation and proposed potential techniques to address this problem \cite{ng1997preventing,prechelt1998automatic}. Here, we applied a simple and straight-forward hyperparameter selection approach to reduce classifier variance. For a given set of hyperparameter $\{S, Q, SFC_{order}\}$, after cross validation with different $C$ values of the linear SVM, we compute the mean and standard deviation to get the $3\sigma$ lower bound over all  $C$ by 

\begin{eqnarray}
score_{3\sigma} = mean(Accuracy) - 3 \times STD(Accuracy) 
\end{eqnarray}  

By selecting the other hyperparameters $\{S, Q, SFC-order\}$ with the best statistical lower bound on the classifier performance over $C$, the optimal hyperparameters have lower variance while preserving lower bias.  Using this hyperparameter selection approach, the classification results are reported in Table \ref{tab:trajresults}.

\begin{table}[t]
  \centering
  \caption{Classification accuracy for TB-Only, RB-TB methods,  multiple normal distribution based symbolic distance (NDist) and our algorithm (\%).}
    \begin{tabular}{lllll}
    \toprule
    Dataset & TB-Only & RB-TB & NDist & GAF-MTF \\
    \midrule
    Animal Tracking & 50 & 83.3 & 83.3 & 72.2 \\
    Hurricane & 65.4 & 73.1 & 52.3 & 71.42 \\
    \bottomrule
    \end{tabular}%
  \label{tab:trajresults}%
\end{table}%

We perform better than the TB-Only method on both datasets and almost as good as the RB-TB method on the 'Hurricane' dataset. However, both RB-TB and NDist methods outperform ours on the 'Animal' dataset.  As shown in Figure \ref{fig:trajdata}, both region and trajectory based features are useful for classification. For the 'Hurricane' dataset, direction based features are more useful than region based features. Direction based features are quite easy to capture using our approach as the GAF is actually calculating the pairwise direction fields on each points in the trajectory data. For the 'Animal' dataset, region is very important as shown in Figure \ref{fig:trajdata} (a). Elk, deer and cattle are almost separable just using location as their regions are clearly located at the left, right top and right bottom, respectively. When transforming the trajectory data into time series using SFC, two close regions might be mapped to different sub-areas with different SFC indexes. When the indexes of two close regions are also near, this can be handled by CNNs with its capability to capture the small shifting-invariance features. However, CNNs are not good at discriminating similar images with large shifting from each other. Thus, when the region information is preserved by the manner of shifting the specific patterns largely in the time series produced by SFC, CNNs might have difficulty capturing the region information.      

Although our approach does not overtake other benchmark methods on both trajectory datasets, we provide a more general framework to encode the spatial-temporal patterns for classification tasks. Instead of complicated hand-tuned features, our approach can be applied to a variety of time series and trajectory data. When the region of the trajectory is not significantly important or the direction feature dominates, our general methods work quite well. On large datasets where the volume of time series/trajectory data is big, our deep neural network based approach will greatly benefit from the large sample size in both feature learning and classification tasks.     

\section{Conclusions and Future Work}

This paper proposed an off-line approach to spatially encode the temporal patterns for classification using convolutional neural networks. We created a pipeline for converting trajectory and time series data into novel
representations, GAF and MTF images, and extracted high-level features
from these using CNNs. The features were subsequently used for
classification. We demonstrated that our approach yields competitive
results when compared to state-of-the-art methods by searching a
relatively small parameter space. We found that GAF-MTF multi-channel
images are scalable to larger numbers of quasi-orthogonal features
that yield more comprehensive images. Our analysis of high-level
features learned from CNNs suggested Tiled CNNs work like
multi-frequency moving averages that benefit from the 2D temporal
dependency that is preserved by the Gramian matrix.  

Important future work will involve applying our method to massive amounts of data and searching in a more complete parameter space to solve real world problems. We are also quite interested in how different deep learning architectures perform on the GAF and MTF images generated from large datasets. Another interesting future direction is to model time series through GAF and MTF images. We aim to apply learned time series models in regression/imputation and anomaly detection tasks. To extend our methods to the streaming data, we suppose to design the online learning approach with recurrent network structures to represent, learn and model temporal data in real-time. 

\section*{References}

\bibliography{mybibfile}

\newpage
\section{Author Biography}
\subsection{Author I}
\begin{wrapfigure}{r}{0.5\textwidth}
  \vspace{-20pt}
  \begin{center}
    \includegraphics[width=0.4\textwidth]{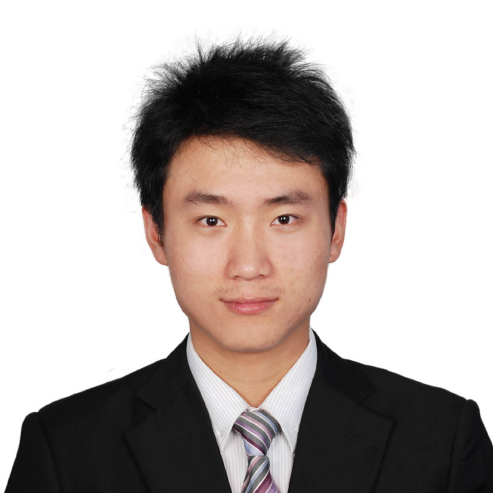}
  \end{center}
  \vspace{-20pt}
\end{wrapfigure}
\noindent {\bf Zhiguang Wang} received his B.S. degree in Mathematics and Applied Mathematics from Fudan University, Shanghai, China, in 2012 and enrolled in the Ph.D. program after then. He received the first prize in Chinese National Mathematical Olympiad in Senior in 2007 and NSF travel award for IJCAI in 2015. He is currently a Ph.D. Candidate in the Department of Computer Science and Electrical Engineering at the University of Maryland Baltimore County. His research interests are in the areas of artificial intelligence, machine learning theory, deep neural networks, non-convex optimization with emphasis on mathematical modeling, time series analysis and pattern recognition in streaming data.

\subsection{Author II}
\begin{wrapfigure}{r}{0.5\textwidth}
  \vspace{-20pt}
  \begin{center}
    \includegraphics[width=0.48\textwidth]{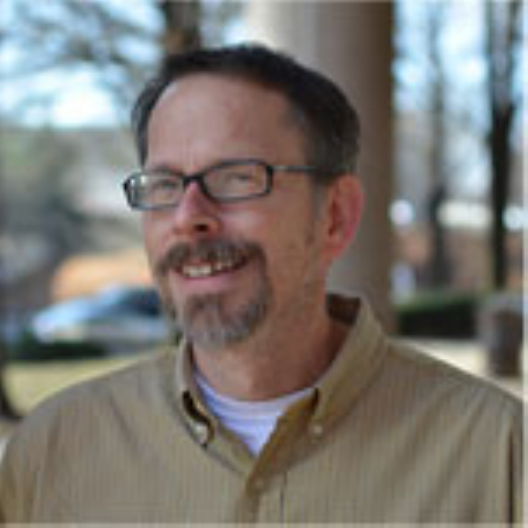}
  \end{center}
  \vspace{-20pt}
\end{wrapfigure}
\noindent {\bf Tim Oates} is a Professor of Computer Science at the University of Maryland Baltimore County. He received B.S. degrees in Computer Science and Electrical Engineering from North
Carolina State University in 1989, and M.S. and Ph.D. degrees from the University of Massachusetts Amherst in 1997 and 2000, respectively. Prior to coming to UMBC in the Fall of 2001, he spent a year as a postdoc in the Artificial Intelligence Lab at the Massachusetts Institute of Technology. In 2004 Dr. Oates won a prestigious NSF CAREER award. He is an author or co-author of more than 190 peer reviewed papers and is a member of the Association for Computing Machinery and the Association for the Advancement of Artificial Intelligence. His research interests include pattern discovery in time series, grammatical inference, graph mining, statistical natural language processing, robotics, and language acquisition.

\end{document}